\documentclass{article}
\usepackage{spconf,amsmath,graphicx}

\usepackage{enumitem}
\setlist{nosep, leftmargin=14pt}

\usepackage{graphicx}
\usepackage{amsmath}
\usepackage{amsfonts}
\usepackage{multirow}
\usepackage{booktabs} 
\usepackage{bbm}
\usepackage{url}
\usepackage{hyperref}
\usepackage{adjustbox}

\title{Baseline Method of the Foundation Model Challenge for Ultrasound Image Analysis}

\name{Bo Deng$^{1}$, Yitong Tang$^{1}$, Jiake Li$^{1}$, Yuxin Huang$^{2}$, Li Wang$^{3,*}$, Yu Zhang$^{1}$, Yufei Zhan$^{1}$, Hua Lu$^{1}$, Xiaoshen Zhang$^{1}$, Jieyun Bai$^{1,*}$}
\address{$^{1}$The First Affiliated Hospital of Jinan University, Jinan University, Guangzhou, China\\
$^{2}$Zhujiang Hospital, Southern Medical University, Guangzhou, China\\
$^{3}$Guangzhou Women and Children's Medical Center, Guangzhou, China\\
$^{*}$Corresponding author: wangli-1227@163.com; jbai996@aucklanduni.ac.nz}

\begin{document}
%
\maketitle

\begin{abstract}
Ultrasound (US) imaging exhibits substantial heterogeneity across anatomical structures and acquisition protocols, posing significant challenges to the development of generalizable analysis models. Most existing methods are task-specific, limiting their suitability as clinically deployable foundation models. To address this limitation, the Foundation Model Challenge for Ultrasound Image Analysis (FM\_UIA~2026) introduces a large-scale multi-task benchmark comprising 27 subtasks across segmentation, classification, detection, and regression. In this paper, we present the official baseline for FM\_UIA~2026 based on a unified Multi-Head Multi-Task Learning (MH-MTL) framework that supports all tasks within a single shared network. The model employs an ImageNet-pretrained EfficientNet--B4 backbone for robust feature extraction, combined with a Feature Pyramid Network (FPN) to capture multi-scale contextual information. A task-specific routing strategy enables global tasks to leverage high-level semantic features, while dense prediction tasks exploit spatially detailed FPN representations. Training incorporates a composite loss with task-adaptive learning rate scaling and a cosine annealing schedule. Validation results demonstrate the feasibility and robustness of this unified design, establishing a strong and extensible baseline for ultrasound foundation model research. The code and dataset are publicly available at \href{https://github.com/lijiake2408/Foundation-Model-Challenge-for-Ultrasound-Image-Analysis}{GitHub}.
\end{abstract}

\begin{keywords}
Ultrasound Analysis, Foundation Model, Multi-task Learning, EfficientNet, Segmentation, Classification, Detection, Regression
\end{keywords}
\section{Introduction}
\label{sec:intro}

Ultrasound (US) imaging presents unique computational challenges due to speckle noise and operator variability, limiting the scalability of traditional task-specific ``Narrow AI'' solutions \cite{liu2019deep}, which are typically restricted to specific benchmarks like fetal head and pubic symphysis segmentation \cite{bai2025psfhs, chen2024psfhs}. Recently, the success of Foundation Models in computer vision, such as the Segment Anything Model (SAM) \cite{kirillov2023segment}, has inspired the development of Generalist Medical AI \cite{moor2023foundation}. A true ultrasound foundation model aims to transcend organ boundaries, handling segmentation, classification, detection, and regression within a single unified framework.

However, training a single network for such heterogeneous tasks is non-trivial. While Multi-Task Learning (MTL) theoretically improves generalization by using related tasks as an inductive bias \cite{caruana1997multitask}, deep networks often suffer from ``negative transfer'' in practice, where optimizing for one task degrades features required for another \cite{vandenhende2021multi}. Balancing these conflicting objectives remains an open research question.

To address this, the Foundation Model Challenge for Ultrasound Image Analysis (FMC\_UIA) 2026 introduces a large-scale benchmark encompassing 27 diverse subtasks. In this paper, we present the official baseline methodology: a Multi-Head MTL framework. We utilize an EfficientNet-B4 encoder with a task-specific routing mechanism to balance the trade-off between semantic abstraction and spatial resolution, providing a robust reference for the competition.

\section{METHOD}
\label{sec:method}

In this section, we delineate the proposed Multi-Head Multi-Task Learning (MH-MTL) framework.

\begin{figure}[t]
\centering
\fboxsep=0pt  
\fbox{\includegraphics[width=3.2 in]{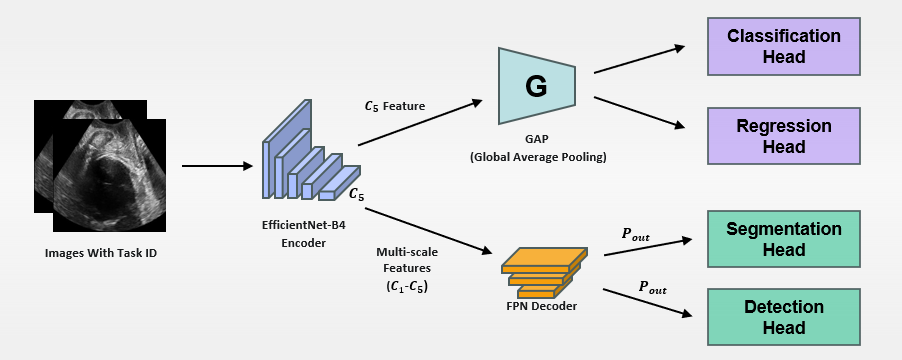}} 
\caption{Overview of the proposed baseline framework. The model utilizes an EfficientNet-B4 backbone for feature extraction. Conditioned on the input task identifier, a task-specific routing mechanism directs the high-level semantic features ($C_5$) to the Global Average Pooling (GAP) module for global tasks (Classification, Regression) or aggregates multi-scale features via the Feature Pyramid Network (FPN) decoder to generate high-resolution feature maps $P_{out}$ for dense tasks (Segmentation, Detection).}
\label{fig:arch}
\end{figure}

\subsection{Unified Network Architecture}
The framework, denoted as $\Phi$, accepts an ultrasound image $X$ and a task identifier $t$. To balance computational efficiency with the requirement for multi-scale feature extraction, all input images are resized to a fixed resolution of $H \times W = 256 \times 256$. The architecture consists of three core components: a shared encoder, a feature pyramid decoder, and a task-routing head module. The overall architecture is illustrated in Fig. \ref{fig:arch}.

\subsubsection{Shared Encoder}
We employ EfficientNet-B4 \cite{tan2019efficientnet} initialized with ImageNet weights as the shared backbone. This encoder extracts a hierarchy of feature maps $\mathcal{F}_{enc} = \{C_1, C_2, C_3, C_4, C_5\}$, where $C_i$ denotes the feature map at stride $2^i$. The deepest layer $C_5$ encapsulates abstract semantic information crucial for diagnostic tasks, while shallower layers retain texture details necessary for boundary delineation.

\subsubsection{Feature Pyramid Decoder}
To address the scale variation inherent in fetal organs and lesions, we incorporate a Feature Pyramid Network (FPN) \cite{lin2017feature}. The FPN aggregates features via top-down pathways and lateral connections, producing a fused multi-scale representation. For dense prediction tasks, we utilize the fused output feature map $P_{out} \in \mathbb{R}^{\frac{H}{4} \times \frac{W}{4} \times D}$, where $D=128$ is the channel dimension.

\subsection{Task-Specific Routing and Heads}
Unlike dynamic routing mechanisms that infer paths at inference time, we implement an explicit Task-Specific Routing strategy. Based on the task type $t$, the information flow is directed to either a Global Branch or a Dense Branch. Specifically, this is implemented via deterministic control flow (i.e., logical branches), where $t$ acts as a switch to activate only the relevant computational path.

\subsubsection{Global Tasks (Classification \& Regression)}
For tasks requiring image-level reasoning (e.g., standard plane classification) or coordinate estimation (e.g., fetal biometry), we utilize the Global Branch. To minimize spatial redundancy, this branch operates solely on the highest-level feature $C_5$. 

The features are processed via Global Average Pooling (GAP), followed by a Dropout layer (rate=0.2) to mitigate overfitting, and a final fully connected projection. The prediction $\hat{Y}_{global}$ is formulated as:
\begin{equation}
    \hat{Y}_{global} = W_{head} \cdot \text{Dropout}(\text{GAP}(C_5)) + b_{head}
\end{equation}

\noindent\textbf{Classification:} The output dimension corresponds to the number of classes $K$, and the logits are processed via Softmax to obtain probabilities.

\noindent\textbf{Regression:} The head outputs a vector of size $2M$ (where $M$ is the number of keypoints), representing normalized $(x, y)$ coordinates.

\subsubsection{Dense Tasks (Segmentation \& Detection)}
Tasks requiring pixel-level localization leverage the FPN output $P_{out}$.

\noindent\textbf{Segmentation:} We employ a convolutional head to project $P_{out}$ into $K$ class masks, followed by upsampling to the original resolution $(H, W)$.

\noindent\textbf{Detection:} To unify detection into this fully convolutional framework, we adopt a simplified anchor-free grid approach. The image is divided into a grid of size $h' \times w'$ (matching the resolution of $P_{out}$). For an object with normalized ground truth center $(x_{gt}, y_{gt})$, we map it to the grid cell $(i, j)$:
\begin{equation}
    i = \lfloor y_{gt} \cdot h' \rfloor, \quad j = \lfloor x_{gt} \cdot w' \rfloor
\end{equation}
The network predicts a vector $\hat{v}_{i,j} \in \mathbb{R}^5$ at each grid cell, consisting of the bounding box coordinates and an objectness score.

\subsection{Objective Functions}
The model is trained using a composite loss function determined by the active task in the current batch.

\subsubsection{Segmentation Loss}
To handle class imbalance between small lesions and the background, we utilize the Dice Loss:
\begin{equation}
    \mathcal{L}_{seg} = 1 - \frac{2 \sum_{p} y_p \hat{y}_p + \epsilon}{\sum_{p} y_p + \sum_{p} \hat{y}_p + \epsilon}
\end{equation}
where $y_p$ and $\hat{y}_p$ are the ground truth and predicted probabilities for pixel $p$.

\subsubsection{Classification Loss}
For diagnostic classification, we employ the standard Cross-Entropy Loss:
\begin{equation}
    \mathcal{L}_{cls} = - \sum_{c=1}^{K} y_c \log(\hat{p}_c)
\end{equation}

\subsubsection{Regression Loss}
For biometric measurements, we calculate the Mean Squared Error (MSE) between the predicted normalized coordinates and the ground truth keypoints:
\begin{equation}
    \mathcal{L}_{reg} = \frac{1}{M} \sum_{k=1}^{M} \| \hat{y}_k - y_k \|_2^2
\end{equation}

\subsubsection{Detection Loss}
Given the sparsity of detection targets in ultrasound (often a single lesion per image), we adopt a focused supervision strategy. We apply the loss specifically at the grid cell $(i, j)$ corresponding to the target center, defined as:
\begin{equation}
    \mathcal{L}_{det} = \mathbb{I}^{\text{obj}}_{i,j} \left[ \mathcal{L}_{bce}(\hat{s}_{i,j}, 1) + \lambda \mathcal{L}_{1}(\hat{b}_{i,j}, b_{gt}) \right]
\end{equation}
where $\mathbb{I}^{\text{obj}}_{i,j}$ is an indicator function that is 1 only at the ground truth center grid and 0 otherwise. $\hat{s}$ represents the objectness score, and $\hat{b}$ represents the bounding box coordinates. $\lambda$ is a balancing hyperparameter set to 8, prioritizing precise localization. This formulation aligns with our baseline implementation, prioritizing capacity on positive samples.

\section{EXPERIMENTS AND RESULTS}
\label{sec:exp}

\subsection{Datasets}
The FMC\_UIA 2026 challenge introduces a large-scale benchmark comprising tens of thousands of ultrasound images derived from multi-center clinical cohorts\cite{lu2022jnu,chen2025comt,zhang2025automatic,chen2024psfhs}. This scale allows for the development of robust generalist models. The dataset exhibits significant heterogeneity in image quality, anatomical views, and acquisition devices, covering 27 distinct subtasks.

For this baseline study, the model was trained on the large-scale training set, and we report quantitative evaluations on the official validation set. This validation set is specifically composed of data from unseen domains to rigorously test the model's generalization capabilities across the four primary task categories:

\begin{itemize}
    \item \textbf{Segmentation (12 subtasks):} Involves pixel-level delineation of fetal organs (e.g., head, heart, abdomen), maternal structures, and lesions. The validation set includes 2,674 samples for these tasks.
    \item \textbf{Classification (9 subtasks):} Covers diagnostic tasks such as fetal standard plane verification and tumor malignancy assessment. The validation set contains 2,727 samples.
    \item \textbf{Detection (3 subtasks):} Focuses on localizing thyroid nodules, uterine fibroids, and spinal cord injuries, with 725 validation samples.
    \item \textbf{Regression (3 subtasks):} Targets biometric measurements including Intra-Uterine Growth Restriction (IUGC) markers and fetal femur length, represented by 617 validation samples.
\end{itemize}

\subsection{Evaluation Metrics}
Performance is evaluated using task-specific metrics standardized by the challenge platform:
\begin{itemize}
    \item \textbf{Segmentation:} We report the Dice Similarity Coefficient (DSC) for volumetric overlap and Hausdorff Distance (HD) for boundary precision.
    \item \textbf{Classification:} To rigorously assess discriminative ability under class imbalance, we utilize the Area Under the Curve (AUC), F1-Score, and Matthews Correlation Coefficient (MCC).
    \item \textbf{Detection:} The Intersection over Union (IoU) measures the localization accuracy of predicted bounding boxes.
    \item \textbf{Regression:} We report the Mean Radial Error (MRE) in pixels. Crucially, MRE is calculated on the original image resolution to reflect real-world clinical measurement precision, rather than on the resized inputs used for training.
\end{itemize}

\subsection{Implementation Details}
The framework is implemented in PyTorch using the Segmentation Models PyTorch (SMP) library. All input images are resized to $256 \times 256$. We apply rigorous data augmentation using the Albumentations library, including Random Brightness Contrast and Gaussian Noise. The model is trained for 50 epochs using the AdamW optimizer with a Cosine Annealing scheduler. The backbone learning rate is set to $1e-4$, while task-specific heads use $1e-3$ to accelerate convergence.

\subsection{Results and Analysis}

\begin{table}[t]
\centering
\caption{Quantitative results of the baseline method on the official FMC\_UIA 2026 validation set. The metrics represent the average across all subtasks within each category.}
\label{tab:results}
\begin{tabular}{l|c|c}
\hline
\textbf{Task Category} & \textbf{Metric} & \textbf{Value} \\
\hline
\hline
\multirow{2}{*}{Segmentation} & Mean DSC ($\uparrow$) & 0.7543 \\
 & Mean HD ($\downarrow$) & 81.18 \\
\hline
\multirow{3}{*}{Classification} & Mean AUC ($\uparrow$) & 0.9155 \\
 & Mean F1-Score ($\uparrow$) & 0.7896 \\
 & Mean MCC ($\uparrow$) & 0.6766 \\
\hline
Detection & Mean IoU ($\uparrow$) & 0.2641 \\
\hline
Regression & Mean MRE (pixels) ($\downarrow$) & 67.43 \\
\hline
\end{tabular}
\end{table}

We evaluated the proposed Multi-Head Multi-Task Learning baseline on the official validation set. The quantitative results, averaged by task category, are summarized in Table \ref{tab:results}.

The experimental results validate the feasibility of using a shared encoder for heterogeneous ultrasound tasks. The model achieves robust performance on Classification (Mean AUC 0.9155) and Segmentation (Mean DSC 0.7543). This indicates that the EfficientNet backbone, initialized with pre-trained weights, effectively captures high-level semantic representations required for distinguishing anatomical planes and delineating major organs. The architecture proves sufficient for tasks where global context or large structural features are dominant.

However, significant performance gaps are observed in localization-sensitive tasks, highlighting the limitations of this unified baseline. Detection yields a modest Mean IoU of 0.2641. This lower performance is primarily observed in subtasks involving small lesions or subtle anomalies. The simplified grid-based detection head, while architecturally unified, struggles to precisely localize small targets compared to large anatomical structures. Additionally, the extreme class imbalance between the background and small lesions in full-sized ultrasound images further impedes convergence.

Similarly, the Regression task shows a Mean MRE of 67.43 pixels. It is important to note that this error is calculated on the original high-resolution images, whereas the model operates on inputs resized to $256 \times 256$. This resolution mismatch inherently limits the spatial precision required for fine-grained biometric measurements.

\section{CONCLUSION}
In summary, while the proposed framework establishes a solid baseline for semantic understanding, the results emphasize the need for future improvements in multi-scale processing and specialized localization modules to address small targets and precise measurements.

\section{Acknowledgments}

This study was supported by project grants from the Guangzhou Municipal Science and Technology Bureau (Grant Nos. 2025B03J0127 and 2023A03J0565) and the Department of Science and Technology of Guangdong Province (Grant No. 2025B1111030001).

\bibliographystyle{IEEEbib}

\end{document}